%% file: acl_latex.tex
\pdfoutput=1

\documentclass[11pt]{article}

\usepackage[preprint]{acl}
\usepackage{hyperref}
\usepackage{times}
\usepackage{latexsym}
\usepackage{amsfonts}  

\usepackage[T1]{fontenc}
\usepackage[utf8]{inputenc}

\usepackage{microtype}

\usepackage{inconsolata}

\usepackage{graphicx}
\usepackage[most]{tcolorbox}
\usepackage{arydshln}
\usepackage{booktabs}
\usepackage{placeins}
\usepackage{tabularx} 
\usepackage{multirow} 
\usepackage{caption}  
\usepackage{array}     
\usepackage{tikz}
\usepackage{float}
\usetikzlibrary{positioning, shapes.geometric, calc}
\usepackage{enumitem}

\FloatBarrier

\title{One Arrow, Many Targets: Probing LLMs for Multi-Attribute Controllable Text Summarization}

\author{Tathagato Roy\textsuperscript{1} \enspace \enspace Rahul Mishra\textsuperscript{2}\\
IIIT Hyderabad\textsuperscript{1}\\
{\tt tathagato.roy@research.iiit.ac.in}, {\tt rahul.mishra@iiit.ac.in}
}

\begin{document}
\maketitle
\begin{abstract}
Text summarization is a well-established task within the natural language processing (NLP) community. However, the focus on controllable summarization tailored to user requirements is gaining traction only recently. While several efforts explore controllability in text summarization, the investigation of Multi-Attribute Controllable Summarization (MACS) remains limited. This work addresses this gap by examining the MACS task through the lens of large language models (LLMs), using various learning paradigms, particularly low-rank adapters. We experiment with different popular adapter fine-tuning strategies to assess the effectiveness of the resulting models in retaining cues and patterns associated with multiple controllable attributes. Additionally, we propose and evaluate a novel hierarchical adapter fusion technique to integrate learnings from two distinct controllable attributes. Subsquently, we present our findings, discuss the challenges encountered, and suggest potential avenues for advancing the MACS task.
\end{abstract}

\section{Introduction}
    \input{latex/introduction}

\section{Related Works}

\input{latex/related_works}

\section{Experimental Setup}

\input{latex/methods}

\section{Results}
    \input{results}
\section{Discussion}
    \input{discussion}

\section{Conclusion}
    \input{conclusion}
\section{Limitations}

\input{limitations}
\section{Ethics Statement}
    \input{latex/ethics}
\FloatBarrier

\section*{Acknowledgments}

For assistance in writing, we utilized large language models (LLMs) such as ChatGPT \footnote{\url{https://openai.com/index/chatgpt/}} and Claude \footnote{\url{https://claude.ai/}}. Their contributions were primarily focused on rephrasing sentences for improved clarity and conciseness.

Furthermore, these LLMs, along with GitHub Copilot \footnote{\url{https://github.com/features/copilot}} were employed to a limited extent for tasks related to LaTeX formatting and occasional code writing, including debugging.

\bibliography{latex/acl_latex}

\appendix

\section{Appendix}
\label{sec:appendix}
\section{Prompts}
\subsection{Prompts Used for Multi-Attribute Controllability}
\label{sec:prompt_MACS}

In this section, we provide some sample prompts used in the experiments for multi-attribute controllability. These prompts are designed to guide the assistant in generating responses based on attributes such as extractiveness, specificity, length, and topic focus. The placeholders \{input\_text\} represent the input text provided to the model.

\subsubsection{Prompt 1: Extractiveness : Normal}
\label{prm}
\begin{quote}
    \textit{You are an honest and to the point assistant, please follow the instruction and answer to the point. Please do not provide any irrelevant information or add any extra words than that is necessary to answer the question. Write a summary of the source text. The summary should be normal in extractiveness. Extractiveness is defined by the degree of exact copying from the source text. The source text is given below.} \{input\_text\}
\end{quote}
\subsubsection{Prompt 2: Specificity : Normal}
\begin{quote}
    \textit{You are an honest and to the point assistant, please follow the instruction and answer to the point. Please do not provide any irrelevant information or add any extra words than that is necessary to answer the question. Write a summary of the source text. The summary should be normal in specificity. Specificity is defined by the degree of detail in the summary. The source text is given below.} \{input\_text\}

\subsubsection{Prompt 3: Extractiveness : Normal and Specificity : Normal}
\begin{quote}
    \textit{You are an honest and to the point assistant, please follow the instruction and answer to the point. Please do not provide any irrelevant information or add any extra words than that is necessary to answer the question. Write a summary of the source text. The summary should be normal in extractiveness. Extractiveness is defined by the degree of exact copying from the source text. The summary should be normal in specificity. Specificity is defined by the degree of detail in the summary. The source text is given below.} \{input\_text\}
\end{quote}

\end{quote}
\subsubsection{Prompt 5: Length: Short}
\begin{quote}
    \textit{You are an honest and to the point assistant, please follow the instruction and answer to the point. Please do not provide any irrelevant information or add any extra words than that is necessary to answer the question. Write a summary of the source text. The summary should be short in length. The length is defined in terms of number of words used in the summary. The source text is given below.} \{input\_text\}
\end{quote}

\subsection{Prompt 6: Topic : (Nepal, Route)}
\begin{quote}
    \textit{You are an honest and to the point assistant, please follow the instruction and answer to the point. Please do not provide any irrelevant information or add any extra words than that is necessary to answer the question. Write a summary of the source text. The summary should be focussed on the topic Nepal, route. The source text is given below.} \{input\_text\}
\end{quote}
\subsection{Prompt 4: Length : Short and Topic : (Nepal, Route)}
\begin{quote}
    \textit{You are an honest and to the point assistant, please follow the instruction and answer to the point. Please do not provide any irrelevant information or add any extra words than that is necessary to answer the question. Write a summary of the source text. The summary should be short in length. The length is defined in terms of number of words used in the summary. The summary should be focussed on the topic Nepal, route. The source text is given below.} \{input\_text\}
\end{quote}
\section{Appendix: Evaluation Prompt for Topical Coherence}

In this section, we present the detailed evaluation prompt used for rating the topical coherence of summaries generated from news articles. The prompt instructs the evaluator on how to score the summary based on the relevance to the requested topics and the input article.

\subsection{Evaluation Prompt}
\label{sec:Evaluation Prompts}
In this section we show the prompt we used to prompt GPT-4-mini for evaluating topical relevance of our summaries with respect to the user defined topics.

\begin{quote}
You will be given one summary written for a news article along with the news article.

Your task is to rate the summary on one metric, that is topical coherence.

Please make sure you read and understand these instructions carefully. Please keep this document open while reviewing, and refer to it as needed.

\textbf{Evaluation Criteria:}

\textbf{Topical Coherence}: Ensure the summary focuses on the topics (there can be more than one topic) requested by the user. The summary should reflect the content of the article and not introduce made-up information. It should draw conclusions or summarize only the details presented in the input. The summary should not include any unrelated or extraneous information that is not aligned with the topic or input article.

\textbf{Evaluation Steps:}
\begin{enumerate}
    \item \textbf{Read the Input Article}: Understand the main topics and points that should be covered in the summary.
    \item \textbf{Read the topic(s) provided by the user.}
    \item \textbf{Read the Summary}: Carefully go through the summary and assess whether it covers the topic in a meaningful and coherent way.
    \item \textbf{Score the Summary}: 1-5 scale:
    \begin{itemize}
        \item 1: The summary somewhat reflects the topic but contains a significant amount of irrelevant or incorrect information or misses relevant information.
        \item 2: The summary is generally on-topic, but may include minor irrelevant details or miss some key points.
        \item 3: The summary is mostly on-topic, covering the requested topic well with very few irrelevant details.
        \item 4: The summary reflects the requested topic with full coherence and no irrelevant or made-up content.
        \item 5: The summary is perfectly on-topic, coherent, and includes all the key points from the input article and is very crisp and to the point.
    \end{itemize}
    \item Just respond with the one score and nothing else.
\end{enumerate}

\textbf{Example:}
\begin{itemize}
    \item \textbf{Source Text:}
    \{document\}
    \item \textbf{Topics:}
    \{topics\}
    \item \textbf{Summary:}
    \{summary\}
    \item \textbf{Evaluation Form (scores ONLY):}
    \begin{itemize}
        \item \textbf{Topical Coherence:}
    \end{itemize}
\end{itemize}

\end{quote}

\clearpage

\section{Full Results and Analysis}
\subsection{Training Details}
    \textbf{Hardware} : A node containing 4  NVIDIA RTX 6000 was used to perform all experiments 

    \textbf{LoRA Setup}
    For all our experiments we use LoRA with $rank = 32$ and $alpha = 16$
    The model was trained using 4-bit quantization with the "nf4" quantization type and bfloat16 for computation \cite{qlora}. We used the default dropout rate of $0.1$. 

    \textbf{HLoRA Setup}: 
    We set $rank_1 = 32$ and $rank_2 = 16$, while setting $alpa_1 = 16$ and $rank_2 = 8$. The lower values for the lower hierarchy is to ensure the initial layer retains its understanding of the first attribute even after the second attribute training is done.

    To reduce computational burden, we have not experimented with adapter fusion for combinations involving specificity and restricted the HLoRA experiments for LLama model with topic, length, and extractiveness.
\subsection{Single Attribute Results}
\label{sec:Single Attribute Result}
    \input{latex/tables/topic_clean}

\input{latex/tables/specificity_only}

\FloatBarrier
\subsection{Multi Attribute Results}
\label{sec:Multi Attribute Result}

\input{latex/tables/length_and_topic_mistral}

\input{latex/tables/length_and_topic_llama}

\input{latex/tables/length_and_specificity_mistral}
\input{latex/tables/length_and_specificity_llama}

\input{latex/tables/extractiveness_and_topic_mistral}

\input{latex/tables/extractiveness_and_topic_llama}

\input{latex/tables/extractiveness_and_specificity_mistral}

\input{latex/tables/extractiveness_and_specificity_llama}

\input{latex/tables/specificity_and_topic_mistral}

\input{latex/tables/specificity_and_topic_llama}

\end{enumerate}

\end{document}

%% file: latex/introduction.tex
Powerful Large Language Models like GPT-3, GPT-4 \cite{gpt3, gpt4} have been found to perform on par with humans on generic summarization tasks \cite{goyal2022zeroshotnews,liu-etal-2023-revisiting, zhang-etal-2024-benchmarking}. 
Generic Summarization is highly open-ended and subjective, and as such, many possible acceptable summaries exist for a given input; therefore, it becomes difficult to precisely evaluate the qualitative differences between many acceptable summaries \cite{instruction}. In practical settings, such summaries might not be the most useful, as different users have different requirements for different summaries. Hence many recent works have focused on the study of Controllable Text Summarization (CTS)  \cite{hydrasum, survey, macsum, instruction}.
\input{summarization_example}
CTS aims to create summaries that fulfill specific criteria specified by the user, by manipulating conditions on various controllable attributes (CAs). These attributes include factors such as summary length \cite{macsum, hydrasum, length_1}, topic \cite{topic_1, macsum}, or even degree of extractiveness \cite{hydrasum, macsum}. The goal is to generate summaries that not only condense the source material but also adhere to particular requirements set by the user for the task at hand.

However, most studies have limited themselves to settings where only one aspect is to be controlled. However, users might often require models to control multiple attributes simultaneously. For example, a user might request "a \textbf{short} summary focusing on \textbf{Lionel Messi}." Hence, we study the LLM's ability to generate summaries that satisfy multiple controllability requirements. This is a hard challenge as LLM's have to focus on what might often be orthogonal or even conflicting requirements. To this end, we explore the ability of existing LLMS on this task of "Multi-Attribute Controllable Summarization(MACS)" in the zero-shot setting and explore various parameter-efficient fine-tuning strategies that allow these models to learn to do this task effectively in the following two settings: 1. Where we have access to labeled data triplets \textbf{(controllable aspect-value pair, input text, human-labeled summaries)} for \textbf{multiple attributes together} and 2. where we have access to such datasets for \textbf{individual attributes independently}. We investigate the trade-off involving using a dataset where we have labeled summaries for multiple attributes jointly vs. combining individual attribute-specific datasets together under a variety of experimental setups.

We leverage recent advances in the \textbf{Parameter Efficient Finetuning Techniques(PEFT)}, specifically LoRA \cite{hu2022lora} to keep the computational cost of such fine-tuning minimal. 

We primarily investigate two research questions 
\begin{enumerate}
    \item \textbf{How effectively do modern Large Language Models (LLMs) handle multi-attribute controllable summarization?} Specifically, how do they manage the challenges of simultaneously controlling multiple attributes (such as length, extractiveness, and topic) in summarization tasks, especially when these control parameters may conflict with or operate independently of each other?
    \item \textbf{ Can models trained to control attributes like length or topic independently be effectively combined in a joint setting to control both attributes simultaneously?}
\end{enumerate}



%% file: summarization_example.tex
\begin{figure}[t]
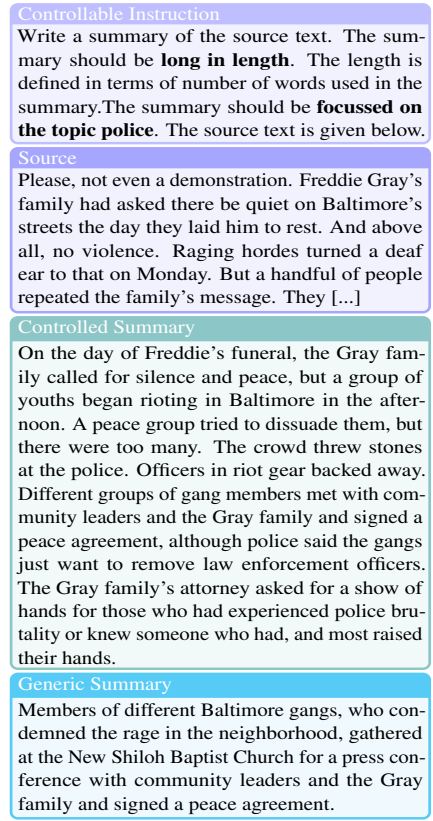

    \setlength{\topsep}{0pt}
    \setlength{\partopsep}{0pt}
    \setlength{\parskip}{0pt}
    \centering
   \scalebox{0.72}[0.65]{ 
    \begin{tcolorbox}[
        colback=blue!5!white,
        colframe=blue!25!white,
        title=Controllable Instruction,
        boxsep=1pt,
        left=2pt,
        right=2pt,
        top=2pt,
        bottom=0pt
    ]
    Write a summary of the source text. The summary should be \textbf{long in length}. The length is defined in terms of number of words used in the summary.The summary should be \textbf{focussed on the topic police}. The source text is given below. 
    \end{tcolorbox}
    }
    \scalebox{0.72}[0.65]{ 
    \begin{tcolorbox}[
        colback=blue!5!white,
        colframe=blue!35!white,
        title=Source,
        boxsep=1pt,
        left=2pt,
        right=2pt,
        top=2pt,
        bottom=1pt
    ]
    Please, not even a demonstration. Freddie Gray's family had asked there be quiet on Baltimore's streets the day they laid him to rest. And above all, no violence. Raging hordes turned a deaf ear to that on Monday. But a handful of people repeated the family's message. They  [...]
    \end{tcolorbox}
    }
    \scalebox{0.72}[0.65]{ 
    \begin{tcolorbox}[
        colback=teal!5!white,
        colframe=teal!45!white,
        title=Controlled Summary,
        boxsep=1pt,
        left=2pt,
        right=2pt,
        top=2pt,
        bottom=1pt
    ]
    On the day of Freddie's funeral, the Gray family called for silence and peace, but a group of youths began rioting in Baltimore in the afternoon. A peace group tried to dissuade them, but there were too many. The crowd threw stones at the police. Officers in riot gear backed away. Different groups of gang members met with community leaders and the Gray family and signed a peace agreement, although police said the gangs just want to remove law enforcement officers. The Gray family's attorney asked for a show of hands for those who had experienced police brutality or knew someone who had, and most raised their hands.
    \end{tcolorbox}
    }
    \scalebox{0.72}[0.65]{ 
    \begin{tcolorbox}[
        colback=cyan!5!white,
        colframe=cyan!55!white,
        title=Generic Summary,
        boxsep=1pt,
        left=2pt,
        right=2pt,
        top=2pt,
        bottom=1pt
    ]
    Members of different Baltimore gangs, who condemned the rage in the neighborhood, gathered at the New Shiloh Baptist Church for a press conference with community leaders and the Gray family and signed a peace agreement.
    \end{tcolorbox}
    }
    \caption{Example of Task of Multi Attribute Controllable Summarization}
    \vspace{-3mm}
\end{figure}

%% file: latex/related_works.tex
\textbf{Controllable Summarization}
As performance of LLMs on summarization benchmark started to reach human performance \cite{goyal2022zeroshotnews,liu-etal-2023-revisiting, zhang-etal-2024-benchmarking}, recent work started to focus on Controllable Text Summarization (CTS). \citet{survey} defines the task of CTS to be of Conditional Generation, conditioned not only on the source documents but also on user-defined instructions in terms of controllable attributes. These attributes can be length \cite{macsum, hydrasum, length_1}, topic \cite{topic_1, macsum}, specificity \cite{macsum, hydrasum}, and even degree of extractiveness \cite{hydrasum, macsum}. Most of the work in these settings is built upon summarization-focused encoder-decoder models like BART \cite{lewis-etal-2020-bart} and Pegasus \cite{pmlr-v119-zhang20ae}. \citet{instruction} benchmarks the ability of LLMs on the closely related task of instruction-controllable summarization. Despite this, there remains a gap in the literature on the effectiveness of LLMs on the task of CTS, especially MACS.

\citet{survey} identifies the area of MACS to be relatively understudied. They identify that prior research has primarily focused on how length and entity attributes interact, but other attributes remain understudied. They notice that while some of these works have methods that work with different attributes, they may not always be able to control multiple attributes jointly. For example \citet{chan-etal-2021-controllable} frames the problem of controllable summarization as a Constrained Markov Decision Process and trains a Reinforcement Learning model to control for length, extractiveness, and entity coverage; we note that these attributes are controlled one at a time and not jointly.
\citet{fan-etal-2018-controllable} trains a simple Convolutional Encoder-Decoder model for generated summaries
to be conditioned on user tokens specifying length, entity, and sources. \citet{hydrasum} trains a mixture of experts' multi-decoder model and observes that different decoders learn to contrast styles and, therefore, strategically sampling from a mixture of decoders allows one to control for multiple attributes effectively. \citet{macsum} is the only work that introduced a dataset for Multi-Attribute CTS where multiple human-written summaries exist for each source conditioned on multiple attributes and benchmarks different baselines on the dataset.

\textbf{Parameter Efficient Finetuning(PEFT)}:
As LLMs and the training dataset become increasingly large, it has become prohibitively expensive to do full fine-tuning of LLMs due to memory(VRAM) and other computational constraints. Hence, a number of methods have been developed to allow fine-tuning of models with only a subset of model parameters \cite{pmlr-v97-houlsby19a, hu2022lora}. LoRA \cite{hu2022lora} and its variants like AdaLoRA \cite{zhang2023adaptive} , DoRA \cite{liu2024doraweightdecomposedlowrankadaptation} have gained widespread adoption due to their straightforward implementation, effectiveness, and adaptability as no underlying model architecture modification is needed and almost all kinds of architectures are supported. QLoRA \cite{qlora} builds on these LoRA variants to support the fine-tuning of quantized models, further reducing the memory footprint of these models.
    
LoRA introduces lower-ranked matrices called adapters, which are fine-tuned instead of the original weights. These adapters can be stored independently of the original weights. This inherently brings about great modularity and composability in how these modules are used in various downstream tasks. 
\citet{ponti2022combining, huang2024lorahub, caccia2023multihead} all explore different strategies to combine different LoRA modules from a library of such independently trained modules for cross-task generalization. There is an interest in the community in better leveraging the inherent composability of these modules and in effectively combining them for intersecting and overlapping tasks.

Pretrained LLMs, trained on unsupervised language modeling and instruction-tuned for tasks like summarization and translation \cite{wei2022finetuned}, have achieved remarkable success. However, they can still misalign with human values, sometimes producing harmful outputs. While extensive instruction tuning can help, it is costly to collect large-scale instruction datasets. Reinforcement Learning from Human Feedback (RLHF) \cite{ouyang2022training} offers an alternative by using human preferences to train a Reward Model, which aligns the model with human preferences. \citet{rafailov2023direct} proposed \textbf{Direct Preference Optimization (DPO)}, a simpler alternative to RLHF, eliminating the need for learning a reward model and complex RL training. 
Since then, many preference alignment methods have been introduced that leverage rankings of outputs annotated by humans to create a preference dataset that can be used to fine-tune the LLMs. Contrastive Preference Optimisation (CPO) \cite{xu2024contrastive} combines maximum-likelihood loss with Direct Preference Optimization (DPO) loss to enhance how models learn and remember preferences. \citet{hong2024orpomonolithicpreferenceoptimization} introduced odds ratio preference optimization(ORPO), which removed the need for reference models and combined preference learning in the SFT phase. 
\input{latex/tables/length_clean}

\input{latex/tables/extractivity}

%% file: latex/tables/length_clean.tex

\begin{table*}[t]
\setlength{\tabcolsep}{4pt}
\centering
\small
\begin{tabular}{l l *{5}{c}}
    \toprule
    \textbf{Model} & \textbf{Finetuning} & \textbf{Metrics} & \textbf{Short} & \textbf{Normal} & \textbf{Long} & \textbf{Overall} \\ 
    \midrule
    \multirow{3}{*}{Llama-3.1-Storm-8B} & \multirow{3}{*}{Zero Shot} 
        & Length          & 46.48 & 104.20 & 196.01 & 112.66\\
        & & Compression Ratio & 0.07 & 0.14 & 0.30 & 0.16 \\
        & & Rouge            & 0.31/0.11/0.22 & 0.28/0.08/0.17 & 0.34/0.11/0.19 & 0.30/0.10/0.19 \\
    \midrule
    \multirow{3}{*}{Mistral-7b-Instruct-v0.3} & \multirow{3}{*}{Zero Shot}
        & Length          & 172.86 & 209.73 & 262.92 & 213.85 \\
        & & Compression Ratio & 0.27 & 0.28 & 0.33 & 0.31 \\
        & & Rouge            & 0.20/0.06/0.13 & 0.22/0.67/0.13 & 0.34/0.12/0.19 & 0.23/0.07/0.14 \\
    \midrule
    \multirow{3}{*}{Llama-3.1-Storm-8B} & \multirow{3}{*}{SFT}
        & Length          & 53.91 & 142.96 & 263.70 & 151.08 \\
        & & Compression Ratio & 0.07 & 0.17 & 0.42 & 0.21 \\
        & & Rouge            & 0.30/0.10/0.22 & 0.25/0.09/0.17 & 0.30/0.12/0.18 & 0.27/0.10/0.19 \\
    \midrule
    \multirow{3}{*}{Mistral-7b-Instruct-v0.3 } & \multirow{3}{*}{SFT}
        & Length          & 37.21 & 64.06 & 91.70 & 64.44 \\
        & & Compression Ratio & 0.06 & 0.08 & 0.15 & 0.09 \\
        & & Rouge            & 0.22/0.06/0.16 & 0.22/0.05/0.14 & 0.25/0.06/0.15 & 0.23/0.06/0.15 \\
    \midrule
    \multirow{3}{*}{Llama-3.1-Storm-8B} & \multirow{3}{*}{DPO}
        & Length          & 39.30 & 100.30 & 296.78 & 132.70 \\
        & & Compression Ratio & 0.06 & 0.13 & 0.46 & 0.19 \\
        & & Rouge            & 0.32/0.12/0.23 & 0.28/0.08/0.18 & 0.30/0.11/0.18 & 0.30/0.10/0.19 \\
    \midrule
    \multirow{3}{*}{Mistral-7b-Instruct-v0.3 } & \multirow{3}{*}{DPO}
        & Length          & 132.58 & 185.80 & 228.18 & 183.64 \\
        & & Compression Ratio & 0.21 & 0.25 & 0.36 & 0.27 \\
        & & Rouge            & 0.22/0.07/0.15 & 0.23/0.07/0.14 & 0.39 & 0.25/0.08/0.15 \\
    \bottomrule
\end{tabular}
\caption{Results on Controlling Length}
\label{tab:length_result}
\vspace{-0.3cm}
\end{table*}

%% file: latex/tables/extractivity.tex
\begin{table*}[t]
\setlength{\tabcolsep}{4pt}
\centering
\small
\begin{tabular}{l l *{5}{c}}
    \toprule
    \textbf{Model} & \textbf{Finetuning} & \textbf{Metrics} & \textbf{Normal} & \textbf{High} & \textbf{Fully} & \textbf{Overall} \\ 
    \midrule
    \multirow{4}{*}{Llama-3.1-Storm-8B} & \multirow{4}{*}{Zero Shot} 
        & Density          & 3.09 & 3.02 & 6.47 & 3.31 \\
        & & Coverage        & 0.90 & 0.91 & 0.92 & 0.90 \\
        & & Overlap Precision         & 0.05 & 0.05 & 0.06 & 0.05 \\
        & & Rouge           & 0.32/0.10/0.20 & 0.23/0.06/0.15 & 0.23/0.07/0.15 & 0.30/0.10/0.19 \\
    \midrule
    \multirow{4}{*}{Mistral-7b-Instruct-v0.3 } & \multirow{4}{*}{Zero Shot} 
        & Density          & 4.76 & 5.16 & 4.82 & 4.79 \\
        & & Coverage        & 0.87 & 0.88 & 0.87 & 0.87 \\
        & & Overlap Precision         & 0.16 & 0.15 & 0.15 & 0.16 \\
        & & Rouge           & 0.24/0.08/0.14 & 0.18/0.07/0.12 & 0.16/0.07/0.12 & 0.23/0.08/0.14 \\
    \midrule
    \multirow{4}{*}{Llama-3.1-Storm-8B} & \multirow{4}{*}{SFT} 
        & Density          & 42.79 & 50.78 & 118.95 & 48.57 \\
        & & Coverage        & 0.94 & 0.93 & 0.98 & 0.95 \\
        & & Overlap Precision        & 0.12 & 0.14 & 0.24 & 0.13 \\
        & & Rouge           & 0.28/0.10/0.18 & 0.20/0.06/0.13 & 0.18/0.07/0.14 & 0.27/0.09/0.18 \\
    \midrule
    \multirow{4}{*}{Mistral-7b-Instruct-v0.3} & \multirow{4}{*}{SFT} 
        & Density          & 3.02 & 2.99 & 2.54 & 2.99 \\
        & & Coverage        & 0.84 & 0.87 & 0.85 & 0.84 \\
        & & Overlap Precision         & 0.03 & 0.02 & 0.02 & 0.03 \\
        & & Rouge           & 0.24/0.06/0.16 & 0.20/0.05/0.13 & 0.19/0.05/0.12 & 0.23/0.06/0.15 \\
    \midrule
    \multirow{4}{*}{Llama-3.1-Storm-8B} & \multirow{4}{*}{DPO} 
        & Density          & 2.95 & 3.08 & 6.69 & 3.21 \\
        & & Coverage        & 0.89 & 0.91 & 0.93 & 0.90 \\
        & & Overlap Precision        & 0.05 & 0.05 & 0.08 & 0.05 \\
        & & Rouge           & 0.32/0.10/0.20 & 0.23/0.05/0.14 & 0.22/0.07/0.15 & 0.30/0.09/0.19 \\
    \midrule
    \multirow{4}{*}{Mistral-7b} & \multirow{4}{*}{DPO} 
        & Density          & 4.78 & 5.21 & 4.34 & 4.78 \\
        & & Coverage        & 0.87 & 0.88 & 0.86 & 0.87 \\
        & & Overlap Precision        & 0.17 & 0.14 & 0.13 & 0.16 \\
        & & Rouge           & 0.23/0.08/0.14 & 0.18/0.07/0.12 & 0.16/0.07/0.12 & 0.22/0.08/0.14 \\
    \bottomrule
\end{tabular}
\caption{Results on controlling only Extractiveness}
\label{tab:extractiveness_result}
\vspace{-0.3cm}
\end{table*}

%% file: latex/methods.tex
We evaluate simple fine-tuning strategies that attempt to adapt the pre-trained LLMs for the task of multi-attribute controllable summarization. For all such strategies, we use parameter-efficient fine-tuning techniques, specifically Low-Rank Adaptation (LoRA) 
for computing, speed, and memory efficiency. We boradly compare two different training objectives 1. \textbf{Supervised Finetuning (SFT)} and 2. \textbf{Direct Preference Optimisation(DPO)} \cite{rafailov2023direct} .

While SFT adapts the LLM to the specific in-task domain represented by examples of (input, output) pair given as supervision, DPO meanwhile provides a more comparative signal where, for a given input text source, the model is shown two different probable output summaries, and the model learns to assign a higher probability to a preferred summary and a lower one to the rejected summary depending on the controllability instruction embedded in the prompt. 

\subsection{Dataset}
\textbf{MACSUM} \cite{macsum} is human labelled dataset used for multi-attribute controllable summarization. They have summaries present for the attributes \textbf{length}, \textbf{extractiveness}, \textbf{topic}, \textbf{specificity} and \textbf{speaker} (for dialogue summarization). We exclude speaker from our experiments. 
\textbf{Length} is defined is terms of the number of words present in the summary. Length is divided is into three categories \textbf{[short, normal, long]}. \textbf{Extractiveness} is the degree to which the summary is extracted/copied from the source text. It is categorized as \textbf{[normal, high, full]}. \textbf{Topic} ensures summaries should only focus on content from certain topics. Finally, \textbf{Specificity} is the degree of detail that the summary should contain. It can take on two values : \textbf{[normal, high]}

\textbf{Preference Dataset Construction} DPO requires the usage of preference data where labeled preferences exist for different outputs given a task. In the absence of such a dataset for our task, we transform \textbf{MACSUM} \cite{macsum} used for supervised fine-tuning to a pairwise preference dataset. In the dataset, for a given input source, we have multiple summaries. For a given input text and controllable aspect, say length, we may have multiple summaries corresponding to the controllable instructions: short, normal and long. Our preference is constructed such that when the controllable instruction is short, the short summary is preferred and the normal and long summaries are the rejected response.

We use 4-bit quantized \textbf{Mistral-7B-Instruct-v0.3}\footnote{mistralai/Mistral-7B-Instruct-v0.1 checkpoint in huggingface} 
and the fine-tuned \textbf{llama3.1-8b-Storm}\footnote{akjindal53244/Llama-3.1-Storm-8B. We encountered challenges in fine-tuning the original Llama model and therefore utilized this checkpoint, which demonstrates comparable performance across various benchmarks. \url{https://huggingface.co/akjindal53244/Llama-3.1-Storm-8B}} models 
as our base models for all our experiment. Our exploratory experiments revealed that smaller 4-bit quantized models, typically within the 1-3 billion parameter range, lack the capacity to perform robust controllable summarization.

We experiment with different configuration of the LORA adapter configuration and training strategies to evaluate the best setup for the task for multi-attribute controllable summarization.

We briefly describe the different fine-tuning strategies used in the following section.

\subsection{Fine-tuning Configurations}
We have limited the scope our study to control two attributes at a time for tractability. 
\begin{enumerate}
    \item \textbf{Single Adapter Continuous}: We use a single LoRA adapter to train first on the dataset with the first attribute and then on the second attribute. Our primary interest in this setup is figuring out whether knowledge gained by the model during the first phase of fine-tuning(for the first attribute) is still retained once fine-tuned on the second attribute
    \item \textbf{Adapter Fusion}: We train two adapters individually on separate attributes and merge them (using weighted linear combination) at test time to evaluate how much of the original adapter information is retained after the merging. There is a lot of work in the existing literature on various ways of merging LoRA adapters 
    \cite{NEURIPS2023_299a08ee, ilharco2023editing}
    We restrict ourselves to simple linear combinations\footnote{We use the \textit{add\_weighted\_adapter} API from Hugging Face: \url{https://huggingface.co/docs/peft/main/en/package_reference/lora}}
.This method has the most promising downstream potential due to its flexibility and cost-effectiveness. 
    \item \textbf{Single Adapter - Jointly trained}: This is the simplest baseline for our task. We fine-tune a LoRA adapter to control multiple attributes simultaneously. While this method is the most straightforward for our task, it is also the most expensive to scale in number of attributes as it requires one to collect labeled data that comprises annotations for multi-attribute controlled summaries. This is not always possible and, hence, restrictive in being adapted for widespread adoption. 
    \item \textbf{Multiple Adapters}: We first train an adapter on attribute one and then freeze the first adapter while training the second adapter on attribute two. Hypothetically, this should improve upon the adapter fusion method as its second attribute is already contextualized by the first attribute and hence should not ideally destructively interfere with the understanding developed regarding the first attribute. Ideally, this should prevent any potential forgetting in the first attribute, as separate adapters are used to retain the information about the different attributes. This also allows greater flexibility over adapter fusion and single adapter methods if we need to focus on one attribute at test time adaptively.

    \item \textbf{Hierarchical LoRA Layers(HLoRA)}: A significant challenge in learning to control two distinct sets of attributes lies in effectively balancing the learning process associated with each. Prolonged fine-tuning can often lead to adapters forgetting previously learned attributes. To address this issue, various strategies have been explored. In our study, we systematically experiment with a novel hierarchical approach to determine whether such a structure is more effective in maintaining balance across these tasks. In LoRA the Weight Matrix $W$ at any given layer is decomposed to lower rank matrices $A$ and $B$. We train on the first attribute with these weights and then freeze them. While fine-tuning on the second attribute we decompose $A$ into two further lower ranked matrices $A_1, A_2$ and $B$ to $B_1, B_2$ and then only train these matrices. 
    The LoRA Forward pass equation is given by :
    $$ h(x) = Wx + BAx $$
    where $W$ is the original weight matrix and $B,A$ are LoRA decomposed lower rank matrices. 
    Our final forward equation after our two stage training is given by :

$$ 
h(x) = Wx + BAx + (B_2 B_1)(A_2 A_1) x 
$$

where \( W \in \mathbb{R}^{N \times M} \), \( A \in \mathbb{R}^{N \times r_1} \), \( B \in \mathbb{R}^{r_1 \times M} \), 
\( B_2 \in \mathbb{R}^{N \times r_2} \), \( B_1 \in \mathbb{R}^{r_2 \times r_1} \), 
\( A_2 \in \mathbb{R}^{r_1 \times r_2} \), and \( A_1 \in \mathbb{R}^{r_2 \times M} \), such that \( r_2 < r_1 \).

\begin{figure}[]  
    \centering
    \includegraphics[width=0.5\textwidth]{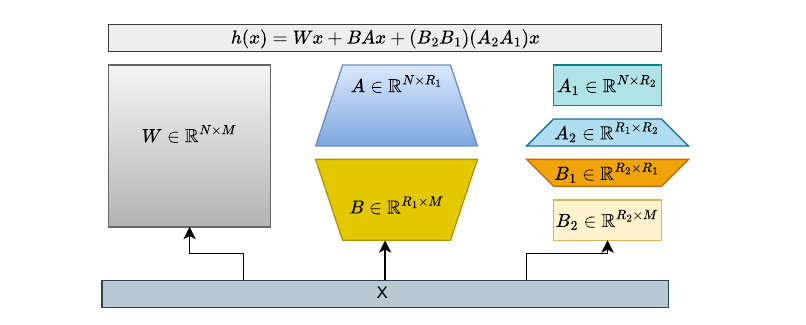}
    \caption{HLoRA Architecture}
    \label{fig:hlora_architecture}  
\end{figure}

\subsection{Metrics}While there exist many standard and widely used metrics to evaluate summaries given a reference summary like ROUGE, Bert-Score \cite{bert-score, lin-2004-rouge}, these metrics evaluate the overall quality of the summary but do not give a sense of to what degree controllability is being achieved. Hence we rely on attribute specific metrics to evaluate the controllability aspect of the summarization along with the standard general quality metric like ROUGE \cite{lin-2004-rouge}

\begin{itemize}[leftmargin=*]
    \item \textbf{Length:} Following \cite{hydrasum}, we report the \textbf{Length} of the summary along with the \textbf{Compression Ratio}, which is defined as ratio of the length of the summary compared to the length of the article

    \item \textbf{Extractiveness:} Following \cite{hydrasum, grusky-etal-2018-newsroom} we report \textbf{Coverage} which represents the proportion of words in the summary that also appear in the input, along with \textbf{Fragmented Density} which measures the average length of consecutive word spans copied from the input into the summary, and \textbf{Overlap Precision} which is the average of 2-gram and 3-gram overlap between the generated summary and the input article. 

    \item \textbf{Topic}: We report the ROUGE \cite{lin-2004-rouge} score. While for other controllable aspects, ROUGE doesn't provide any signal regarding the controllability aspect of the evaluation, for topic, high n-gram level overlap indicated by rouge does signal topical summaries when matched with ground-truth. 
    There is a lack of easily adaptable automatic metric for evaluating topical coherence. \cite{passali-tsoumakas-2024-topic} uses a model trained on a corpus related to given topic 
    to compute similarity between a summary and topic, but such methods cannot be directly adapted as it requires there to be a training document of words for each possible topic which is not always available. Hence, following \cite{instruction, zhang-etal-2024-benchmarking} we employ a strong LLM to evaluate the outputs of our experiments.
    We closely follow the framework of G-eval \cite{liu-etal-2023-g} in prompting GPT-4-mini \cite{gpt4} with Chain-of-Thought (COT) to ask the model to rate the relevance of the generated summary with the prompted topic on a 1-5 Likert Scale. We differ from G-eval in that we report a simple average over a weighted average to reduce output token costs (as computing the token probabilities requires sampling from the model multiple times). Our prompts can be found in the appendix \ref{sec:Evaluation Prompts}. 

    \item \textbf{Specificity} Following \cite{hydrasum}, we report the macro average of sentence level specificity score as given by the \texttt{Speciteller} tool \cite{10.5555/2886521.2886638}.
\end{itemize}

\input{latex/tables/length_and_extractiveness_mistral}

\input{latex/tables/length_and_extractivity_llama}


%% file: latex/tables/length_and_extractiveness_mistral.tex
\begin{table*}[ht]
\setlength{\tabcolsep}{4pt}
\centering
\small
\resizebox{\textwidth}{!}{
\begin{tabular}{|l|l|l|ll|lll|}
\toprule
\multirow{2}{*}{} & \multirow{2}{*}{Method} & \multirow{2}{*}{Config} & \multicolumn{2}{c|}{Length Metrics(S/N/L)} & \multicolumn{3}{c|}{Extractiveness Metrics(N/H/F)} \\
\cmidrule(lr){4-5} \cmidrule(lr){6-8}
& & & Length & Comp. Ratio & Density & Coverage & Overlap Precision \\
\midrule
zero shot &  &  & 169.34/206.98/258.55 & 0.27/0.28/0.4 & 3.92/3.9/4.32 & 0.85/0.87/0.86 & 0.13 / 0.12 / 0.13 \\
\midrule
joint training & SFT &  & 50.74/58.0/85.09 & 0.08/0.07/0.13 & 2.83/2.4/2.64 & 0.84/0.82/0.86 & 0.03 / 0.02 / 0.02 \\
joint training & DPO &  & 152.05/202.25/241.41 & 0.24/0.27/0.38 & 3.94/3.74/3.48 & 0.86/0.86/0.83 & 0.13 / 0.1 / 0.11 \\
\midrule
adapter fusion & SFT & 0.67, 0.33 & 52.39/60.04/98.43 & 0.07/0.08/0.14 & 3.83/6.02/3.18 & 0.85/0.86/0.83 & 0.03 / 0.03 / 0.04 \\
adapter fusion & SFT & 0.5, 0.5 & 50.99/57.88/95.69 & 0.07/0.07/0.15 & 3.03/3.2/5.13 & 0.85/0.86/0.86 & 0.03 / 0.03 / 0.03 \\
adapter fusion & SFT & 0.33, 0.67 & 48.74/61.33/99.23 & 0.07/0.08/0.15 & 3.22/4.26/2.69 & 0.84/0.88/0.85 & 0.04 / 0.02 / 0.03 \\
adapter fusion & DPO & 0.67, 0.33 & 138.38/191.78/237.79 & 0.23/0.26/0.38 & 3.83/4.04/4.44 & 0.86/0.88/0.86 & 0.12 / 0.12 / 0.11 \\
adapter fusion & DPO & 0.5, 0.5 & 147.77/198.6/243.31 & 0.24/0.27/0.38 & 3.97/4.42/3.83 & 0.86/0.87/0.85 & 0.13 / 0.11 / 0.11 \\
adapter fusion & DPO & 0.33, 0.67 & 155.92/204.84/250.11 & 0.25/0.27/0.4 & 3.99/4.03/4.25 & 0.86/0.87/0.86 & 0.13 / 0.11 / 0.12 \\
\midrule
single adapter continuouss & SFT & l->e & 39.22/56.89/84.62 & 0.06/0.07/0.14 & 3.34/2.9/2.74 & 0.84/0.83/0.85 & 0.03 / 0.02 / 0.02 \\
single adapter continuouss & SFT & e->l & 36.78/52.98/93.76 & 0.06/0.07/0.13 & 2.98/3.33/2.93 & 0.83/0.81/0.82 & 0.03 / 0.02 / 0.02 \\
single adapter continuouss & DPO & l->e & 134.7/192.43/242.2 & 0.22/0.26/0.38 & 4.1/4.26/4.27 & 0.86/0.88/0.86 & 0.13 / 0.11 / 0.11 \\
single adapter continuouss & DPO & e->l & 129.55/187.4/245.19 & 0.21/0.26/0.38 & 4.07/3.78/3.94 & 0.86/0.87/0.86 & 0.13 / 0.1 / 0.09 \\
\midrule
multi adapter & SFT & l->e & 40.26/57.24/101.17 & 0.06/0.07/0.17 & 2.75/2.73/2.36 & 0.84/0.86/0.84 & 0.03 / 0.03 / 0.02 \\
multi adapter & SFT & e->l & 45.02/57.07/102.16 & 0.07/0.07/0.16 & 3.45/3.05/3.25 & 0.84/0.84/0.81 & 0.04 / 0.03 / 0.03 \\

multi adapter & DPO & l->e & 164.7/209.73/252.85 & 0.26/0.28/0.4 & 4.2/4.28/3.77 & 0.86/0.88/0.85 & 0.14 / 0.12 / 0.11 \\
multi adapter & DPO & e->l & 137.38/196.49/248.78 & 0.22/0.27/0.39 & 4.35/4.19/3.86 & 0.86/0.88/0.86 & 0.13 / 0.11 / 0.11 \\

\bottomrule
\end{tabular}
}
\caption{
Mistral-7b-Instruct-v0.3 Model results on controlling Length and Extractiveness Metrics.
(N/H/F) refers to Normal, High, Fully for attractiveness and (S/N/L) refers to Short, Normal and Long respectively for length. For metrics presented in this way we care more about the trend(whether it is increasing) than the exact values.
Note that config refers to the weights of the individual attribute adapter when weighted fusion was done for the adapter fusion method, and order of training on the adapters for the rest of the rows where it is not empty. l->e refers to length, then extractiveness, and vice versa for e->l.
Adapter fusion especially while using DPO seems to most promising avenue of jointly controlling length and extractiveness.
}
\label{tab:mistral_length_and_extractiveness}
\end{table*}

%% file: latex/tables/length_and_extractivity_llama.tex
\begin{table*}[ht]
\setlength{\tabcolsep}{4pt}
\centering
\small
\resizebox{\textwidth}{!}{
\begin{tabular}{|l|l|l|ll|lll|}
\toprule
\multirow{2}{*}{} & \multirow{2}{*}{Method} & \multirow{2}{*}{Config} & \multicolumn{2}{c|}{Length Metrics(S/N/L)} & \multicolumn{3}{c|}{Extractiveness Metrics(N/H/F)} \\
\cmidrule(lr){4-5} \cmidrule(lr){6-8}
& & & Length & Comp. Ratio & Density & Coverage & Overlap Precision \\
\midrule
zero shot &  &  & 51.7/112.2/233.97 & 0.08/0.15/0.36 & 3.29/3.28/3.65 & 0.89/0.9/0.89 & 0.08 / 0.06 / 0.08 \\
\midrule
joint training & SFT &  & 48.18/163.06/281.98 & 0.07/0.19/0.45 & 48.93/52.79/111.85 & 0.95/0.95/0.98 & 0.16 / 0.14 / 0.29 \\
joint training & DPO &  & 50.92/122.98/321.18 & 0.08/0.16/0.5 & 3.88/3.82/12.93 & 0.88/0.9/0.91 & 0.1 / 0.07 / 0.12 \\
\midrule

adapter fusion & SFT & 0.67, 0.33 & 63.47/164.22/313.2 & 0.09/0.2/0.51 & 63.04/75.01/59.76 & 0.95/0.95/0.97 & 0.19 / 0.15 / 0.23 \\
adapter fusion & SFT & 0.5, 0.5 & 50.39/181.08/331.32 & 0.08/0.22/0.52 & 70.62/62.7/78.16 & 0.95/0.95/0.97 & 0.21 / 0.15 / 0.23 \\
adapter fusion & SFT & 0.33, 0.67 & 62.35/186.51/311.71 & 0.09/0.23/0.47 & 65.25/63.26/113.97 & 0.95/0.93/0.96 & 0.19 / 0.18 / 0.25 \\
adapter fusion & DPO & 0.67, 0.33 & 45.08/106.65/310.54 & 0.07/0.14/0.49 & 5.58/9.47/10.34 & 0.89/0.9/0.91 & 0.1 / 0.09 / 0.12 \\
adapter fusion & DPO & 0.5, 0.5 & 45.62/109.47/283.59 & 0.07/0.15/0.44 & 4.47/3.53/6.71 & 0.89/0.9/0.91 & 0.09 / 0.07 / 0.1 \\
adapter fusion & DPO & 0.33, 0.67 & 48.49/112.7/263.5 & 0.07/0.15/0.42 & 3.47/3.26/11.79 & 0.89/0.9/0.91 & 0.09 / 0.06 / 0.1 \\

\midrule

single adapter continuous & SFT & l->e & 52.62/196.4/267.67 & 0.07/0.23/0.41 & 83.55/129.58/111.52 & 0.97/0.98/0.99 & 0.19 / 0.22 / 0.27 \\
single adapter continuous & SFT & e->l & 83.41/148.57/228.71 & 0.1/0.18/0.35 & 71.42/71.01/133.92 & 0.97/0.97/0.98 & 0.16 / 0.14 / 0.24 \\
single adapter continuous & DPO & l->e & 39.1/116.03/353.83 & 0.06/0.16/0.58 & 5.34/3.84/17.76 & 0.89/0.9/0.92 & 0.11 / 0.08 / 0.17 \\
single adapter continuous & DPO & e->l & 50.18/94.24/249.74 & 0.08/0.12/0.4 & 3.71/4.2/10.56 & 0.91/0.92/0.95 & 0.08 / 0.06 / 0.15 \\
\midrule

multi adapter & SFT & l->e & 64.5/208.2/334.34 & 0.09/0.25/0.51 & 82.95/73.69/103.35 & 0.95/0.96/0.97 & 0.22 / 0.2 / 0.25 \\
multi adapter & SFT & e->l & 91.76/175.78/301.79 & 0.13/0.2/0.46 & 66.54/68.27/143.12 & 0.96/0.96/0.98 & 0.19 / 0.15 / 0.35 \\
multi adapter & DPO & l->e & 48.5/112.79/244.09 & 0.07/0.15/0.38 & 3.4/3.01/11.9 & 0.88/0.89/0.9 & 0.08 / 0.06 / 0.09 \\
multi adapter & DPO & e->l & 54.02/96.15/247.19 & 0.08/0.12/0.38 & 4.84/3.49/12.57 & 0.91/0.92/0.94 & 0.08 / 0.06 / 0.15 \\

\midrule 
HLoRA & SFT & e->l & 50.34/84.87/112.63 & 0.07/0.1/0.17 & 16.76/34.73/9.6 & 0.96/0.97/0.96 & 0.07 / 0.11 / 0.06 \\
HLoRA & SFT & l->e & 34.54/54.23/91.16 & 0.05/0.07/0.14 & 13.86/11.4/10.48 & 0.95/0.95/0.95 & 0.06 / 0.03 / 0.06 \\

\bottomrule
\end{tabular}
}
\caption{
Llama-3.1-Storm-8B Model results on controlling Length and Extractiveness Metrics.
(N/H/F) refers to Normal, High, Fully for extractivenes and (S/N/L) refers to Short, Normal and Long respectively for length. For metrics presented in this way we care more about the trend(whether it is increasing) than the exact values.
Note that config refers to the weights of the individual attribute adapter when weighted fusion was done for the adapter fusion method, and order of training on the adapters for the rest of the rows where it is not empty. l->e refers to length, then extractiveness, and vice versa for e->l.
Model is able to vary the length as required from the input prompt but is unable to control attractiveness whatsoever.
}
\label{tab:llama_length_and_extractiveness}
\end{table*}


%% file: results.tex
Owing to space limitations, we present a subset of the results here, with the full results available in the Appendix(see \ref{sec:Single Attribute Result} , \ref{sec:Multi Attribute Result}). 
We report the single-attribute results to provide the context along with the multi-attribute results. 

Our analysis shows that the attributes can be divided into categories: easy and hard. 
Length (Table \ref{tab:length_result}) and Topic (Table \ref{tab:topic_result}) fall into the easy categories. Both models exhibited reasonable control over the two specified attributes, even in the zero-shot setting. Fine-tuning largely preserved this control. Furthermore, adding an additional attribute did not significantly compromise their ability to maintain control over the original attributes. We observed that both Mistral and Llama tended to generate summaries that were excessively long compared to human-generated summaries, demonstrating a bias towards comprehensiveness over brevity. Fine-tuning  doesn't seem to change this characteristic significantly. For topic, we are able to determine that DPO improves its performance, while SFT degrades it. We believe that the contrastive nature of DPO better signals to the model what is expected of the model in case of giving topical summaries. 

Specificity (Table \ref{table:specificity_result}) and Extractiveness(Table \ref{tab:extractiveness_result}) can be classified as hard attributes. The model is unable to reliably control these attributes even after fine-tuning. For extractiveness, we notice that when training the Mistral-based model using SFT, the model learns to copy large chunks of the text from source articles, as seen from the fragmented density numbers (see Tables \ref{tab:extractiveness_result} and \ref{tab:llama_length_and_extractiveness}), but is unable to vary this as required from the prompt. 
For the Llama-based model, there is surprisingly some evidence (Table \ref{tab:llama_extractiveness_specificity} that controlling both together using joint training can be done successfully.
Evidence from our experiments strongly suggests that fusing adapters for MACS is often unreliable, particularly when controlling attributes such as specificity or extractiveness. While a jointly labeled dataset does not guarantee effective controllability, it remains the most promising approach.
Furthermore, the optimal control strategy can vary significantly across different models. This variability is likely attributed to the unique post-training dynamics and settings of each model. We have refrained from extensive fine-tuning to preserve the model's original capabilities and avoid overfitting.
This model-dependent behavior presents a challenge for practitioners seeking a universal toolkit. Our findings underscore the necessity of careful handling, prompt engineering, and hyperparameter tuning to achieve desired performance in MACS applications.

%% file: discussion.tex
The control of multiple attributes together poses a significant challenge for Large Language Models (LLMs) despite the sophistication of current models. While certain attributes, such as output length and topical focus, demonstrate relative tractability, others, including extractiveness and specificity, pose significant difficulties.  
Attributes that align closely with quantifiable metrics or well-defined categories in training data, such as subject matter, are more readily manipulated. Conversely, attributes requiring nuanced contextual understanding, like determining appropriate levels of detail or determining the degree of borrowing from the source text, are perhaps more challenging for the models to reliably understand. These challenges are exacerbated by the interdependence of attributes and the complexity of prompt engineering. Consequently, the development of more sophisticated control mechanisms for LLMs, along with more robust ones, remains an active area of research, with implications for enhancing the utility and reliability of these models in various applications.

In-context learning and improved prompting strategies offer promising avenues for enhancing multi-attribute controllability in language models. By leveraging the model's existing knowledge and ability to generalize from examples, in-context learning enables better internalization of attribute relationships without extensive fine-tuning. Carefully crafted prompts can further guide the model to focus on specific aspects of the input, ensuring outputs that align with multiple attribute constraints, such as specificity or extractiveness. We leave the work of exploring the effectiveness of in-context learning and advanced prompting techniques like Chain-of-Thought prompting \cite{10.5555/3600270.3602070} for this task as future work.

DPO tends to work better for controllability, at least for the topic, while also showing more diversity for length compared to Supervised Fine-Tuning (SFT). The models greatly benefit from contrastive signals, making it easier for the model to distinguish between correct and incorrect behaviors. In DPO, the model is presented with pairs of outputs, one preferred and one not, which helps the model learn more effectively by emphasizing what actions to take and, just as importantly, what to avoid. This contrastive feedback is particularly useful for topic control, as it allows the model to clearly identify relevant information and discard off-topic or irrelevant content, leading to more accurate and focused outputs.

%% file: conclusion.tex
Despite the remarkable advancements in large language models (LLMs), they continue to face challenges in multi-attribute controllable summarization, particularly for demanding attributes such as extractiveness and specificity.

Our empirical investigations reveal that combining independent control of individual attributes within a joint framework is not straightforward and necessitates careful experimentation. This highlights the complexity of achieving effective control over multiple aspects of the summarization process.

%% file: limitations.tex
While we are studying multi-attribute controllability, for the sake of tractability, we have limited our study to the case where the number of attributes to control $N = 2$. 

Due to computational limitations, we were compelled to employ quantized models. Specifically, we reduced the model parameters to 4-bit precision. Although this quantization is known to introduce performance degradation, it remains a valuable research avenue, given the prevalence of quantized models in real-world applications due to similar computational constraints. While our study may be subject to limitations arising from quantization, its insights are still relevant and contribute to understanding the behavior of quantized models in practical settings
Our study is also limited to the news domain as the document partition of the MACSUM dataset \cite{macsum} is derived from the CNN-DM dataset \cite{see-etal-2017-get}. Our prompts or our takeaways may not extend to medical or dialogue domains. 

We have also limited our study to attributes for which multi-attribute benchmarks are available. This, therefore, ignores many other essential controllability aspects like style, readability, aspect, etc.

%% file: latex/ethics.tex
This paper evaluates LLMs on multi-attribute controllable summarization tasks. Given the technical and evaluative nature of our research, we do not anticipate any negative ethical or social implications
 

%% file: latex/tables/topic_clean.tex
\begin{table*}[t]
\setlength{\tabcolsep}{4pt}
\centering
\small
\begin{tabular}{ccc|cc}
\toprule
Model & Finetuning & Rouge(1/2/L) & G-eval Score & G-eval Std \\
\midrule
Llama-3.1-Storm-8B & Zero Shot &   0.30 / 0.10 / 0.20 & 4.30 & 0.58 \\
Mistral-7B-Instruct-v0.3 & Zero Shot &   0.16 / 0.06 / 0.11 & 4.45 & 0.64 \\
Llama-3.1-Storm-8B & SFT &   0.28 / 0.09 / 0.19 & 4.21 & 0.63 \\
Mistral-7B-Instruct-v0.3 & SFT   & 0.22 / 0.06 / 0.15 & 3.79 & 0.91 \\
Llama-3.1-Storm-8B & DPO  & 0.31 / 0.10 / 0.20 & 4.23 & 0.65 \\
Mistral-7B-Instruct-v0.3 & DPO   & 0.22 / 0.08 / 0.15 & 4.47 & 0.63 \\

\bottomrule
\end{tabular}
\caption{Results on Controlling Topic. SFT degrades the quality of topical summary especially for Mistral while DPO either maintains the status quo or improves it}
\label{tab:topic_result}
\vspace{-0.3cm}
\end{table*}

%% file: latex/tables/specificity_only.tex
\begin{table*}[t]
\setlength{\tabcolsep}{4pt}
\centering
\small
\begin{tabular}{l l *{4}{c}}
    \toprule
    \textbf{Model} & \textbf{Finetuning} & \textbf{Metrics}& \textbf{Normal} & \textbf{High} & \textbf{Overall} \\ 
    \midrule
 \multirow{2}{*}{Llama-3.1-Storm-8B} & \multirow{2}{*}{Zero Shot} 
        & Specificity          & 0.56 & 0.57 & 0.56 \\
        & & Rouge            & 0.32/0.25/0.31 & 0.10/0.07/0.10 & 0.20/0.16/0.19 \\
    \multirow{2}{*}{Mistral-7B-Instruct-v0.3} & \multirow{2}{*}{Zero Shot} 
        & Specificity          & 0.78 & 0.79 & 0.78 \\
        & & Rouge            & 0.23/0.17/0.23 & 0.08/0.06/0.08 & 0.14/0.12/0.14 \\
    \multirow{2}{*}{Llama-3.1-Storm-8B} & \multirow{2}{*}{SFT} 
        & Specificity          & 0.52 & 0.53 & 0.52 \\
        & & Rouge            & 0.28/0.21/0.27 & 0.10/0.07/0.09 & 0.19/0.14/0.18 \\
    \multirow{2}{*}{Mistral-7B-Instruct-v0.3} & \multirow{2}{*}{SFT} 
        & Specificity          & 0.70 & 0.76 & 0.71 \\
        & & Rouge            & 0.24/0.22/0.24 & 0.06/0.05/0.06 & 0.16/0.15/0.16 \\
    \multirow{2}{*}{Llama-3.1-Storm-8B} & \multirow{2}{*}{DPO} 
        & Specificity          & 0.56 & 0.57 & 0.56 \\
        & & Rouge            & 0.32/0.25/0.31 & 0.10/0.07/0.10 & 0.20/0.16/0.19 \\
    \multirow{2}{*}{Mistral-7B-Instruct-v0.3} & \multirow{2}{*}{DPO} 
        & Specificity          & 0.77 & 0.78 & 0.77 \\
        & & Rouge            & 0.23/0.17/0.22 & 0.08/0.06/0.08 & 0.14/0.12/0.14 \\
    \bottomrule
\end{tabular}
\caption{Results on Controlling Specificity using LoRA Finetuning. Finetuning seems to largely ineffective for controlling specificity. Using Mistral with SFT is the only method which is able to control specificity to an extent
}
\label{table:specificity_result}
\vspace{-0.3cm}
\end{table*}

%% file: latex/tables/length_and_topic_mistral.tex
\begin{table*}[ht]
\setlength{\tabcolsep}{4pt}
\centering
\tiny
\resizebox{\textwidth}{!}{%
\begin{tabular}{|l|l|l|ll|lll|}
\toprule
\multirow{2}{*}{} & \multirow{2}{*}{Method} & \multirow{2}{*}{Config} & \multicolumn{2}{c|}{Length Metrics(S/N/L)} & \multicolumn{3}{c|}{Topic Metrics} \\
\cmidrule(lr){4-5} \cmidrule(lr){6-8}
& & & Length & Comp. Ratio & Rouge(1/2/L) & G-eval Score & G-eval Std\\
\midrule
zero shot &    &  & 169.87/222.70/289.19 & 0.23/0.27/0.36  & 0.18 / 0.06 / 0.12 & 4.42 & 0.66 \\
\midrule

joint training & SFT &   & 55.90/57.79/123.47 & 0.07/0.07/0.14  & 0.21 / 0.05 / 0.14 & 2.55 & 1.17 \\
joint training & DPO  & & 136.52/167.36/217.22 & 0.18/0.20/0.27  & 0.21 / 0.07 / 0.14 & 4.43 & 0.67 \\
\midrule
adapter fusion & SFT & 0.67, 0.33 & 71.00/69.06/100.19 & 0.08/0.07/0.13  & 0.22 / 0.06 / 0.15 & 3.06 & 1.24 \\
adapter fusion & SFT & 0.5, 0.5 & 44.16/61.54/95.53 & 0.06/0.07/0.12  & 0.23 / 0.06 / 0.16 & 3.19 & 1.20 \\
adapter fusion & SFT & 0.33, 0.67 & 78.84/70.69/88.53 & 0.09/0.08/0.11  & 0.22 / 0.06 / 0.15 & 3.29 & 1.12 \\
adapter fusion & DPO & 0.67, 0.33 & 131.97/177.68/239.89 & 0.17/0.21/0.31  & 0.20 / 0.07 / 0.13 & 4.42 & 0.67 \\
adapter fusion & DPO & 0.5, 0.5 & 136.03/173.45/223.86 & 0.18/0.21/0.28  & 0.21 / 0.07 / 0.13 & 4.50 & 0.56 \\
adapter fusion & DPO & 0.33, 0.67 & 124.55/159.81/219.06 & 0.16/0.19/0.28  & 0.21 / 0.07 / 0.14 & 4.44 & 0.62 \\
\midrule

single adapter continuous & SFT & t->l & 33.42/53.07/122.36 & 0.04/0.06/0.16  & 0.21 / 0.05 / 0.14 & 2.46 & 1.19 \\
single adapter continuous & SFT & l->t & 52.77/50.23/94.39 & 0.06/0.06/0.11  & 0.19 / 0.04 / 0.14 & 2.12 & 1.15 \\
single adapter continuous & DPO & t->l & 116.00/173.74/231.31 & 0.15/0.21/0.30  & 0.21 / 0.07 / 0.14 & 4.43 & 0.62 \\
single adapter continuous & DPO & l->t & 81.23/114.91/172.14 & 0.12/0.14/0.21  & 0.24 / 0.08 / 0.16 & 4.35 & 0.70 \\
\midrule

multi adapter & SFT & t->l & 40.71/58.43/103.33 & 0.05/0.06/0.12  & 0.21 / 0.05 / 0.14 & 2.61 & 1.17 \\
multi adapter & SFT & l->t & 61.10/77.36/118.19 & 0.08/0.09/0.15  & 0.23 / 0.06 / 0.16 & 3.57 & 1.02 \\
multi adapter & DPO & t->l & 134.74/190.01/259.69 & 0.18/0.23/0.34  & 0.19 / 0.06 / 0.13 & 4.39 & 0.67 \\
multi adapter & DPO & l->t & 116.52/143.14/203.14 & 0.15/0.17/0.25  & 0.23 / 0.08 / 0.16 & 4.41 & 0.70 \\

\bottomrule
\end{tabular}%
}

\caption{
Mistral-7B-Instruct-v0.3 Model results on controlling Length and Topic
(S/N/H) refers to Short, Normal, High for length  . For metrics presented in this way we care more about the trend(whether it is increasing) rather than the exact values.
 Note that config refers to the weights of the individual attribute adapter when weighted fusion was done for the adapter fusion method, and order of training on the adapters for the rest of the rows, and that t->l refers to topic, then length, and vice versa for l->t.
 GPT-4-mini consistently dislikes topical summaries trained on  SFT while DPO improves the performance on topic controllability. All methods are adequately control length.
}

\label{tab:mistral_length_topic}
\end{table*}

%% file: latex/tables/length_and_topic_llama.tex
\begin{table*}[ht]
\setlength{\tabcolsep}{4pt}
\centering
\tiny
\resizebox{\textwidth}{!}{%
\begin{tabular}{|l|l|l|ll|lll|}
\toprule
\multirow{2}{*}{} & \multirow{2}{*}{Method} & \multirow{2}{*}{Config} & \multicolumn{2}{c|}{Length Metrics(S/N/L)} & \multicolumn{3}{c|}{Topic Metrics} \\
\cmidrule(lr){4-5} \cmidrule(lr){6-8}
& & & Length & Comp. Ratio & Rouge(1/2/L) & G-eval Score & G-eval Std\\
\midrule
zero shot &   &  & 48.35/101.35/212.22 & 0.06/0.12/0.27 &   0.26 / 0.09 / 0.18 & 4.35 & 0.58 \\
\midrule

joint training & SFT &  & 67.10/155.93/245.14 & 0.07/0.17/0.30 &  0.23 / 0.08 / 0.16 & 3.67 & 1.03 \\
joint training & DPO &  & 40.10/85.40/189.06 & 0.05/0.10/0.24 &   0.27 / 0.09 / 0.18 & 4.34 & 0.53 \\
\midrule

adapter fusion & SFT & 0.33, 0.67 & 48.74/108.69/268.31 & 0.06/0.12/0.33 &   0.25 / 0.09 / 0.17 & 4.14 & 0.74 \\
adapter fusion & SFT & 0.67, 0.33 & 58.39/117.47/284.22 & 0.06/0.13/0.36  & 0.25 / 0.09 / 0.18 & 3.89 & 0.93 \\
adapter fusion & DPO & 0.33, 0.67 & 36.74/93.44/256.56 & 0.05/0.11/0.34 &   0.27 / 0.09 / 0.18 & 4.33 & 0.52 \\
adapter fusion & DPO & 0.67, 0.33 & 40.23/94.89/284.86 & 0.05/0.11/0.35 &   0.27 / 0.09 / 0.18 & 4.32 & 0.56 \\
adapter fusion & DPO & 0.5, 0.5 & 41.13/91.85/264.47 & 0.05/0.11/0.34 &   0.27 / 0.09 / 0.18 & 4.29 & 0.55 \\
\midrule

single adapter continuous & SFT & t->l & 60.48/144.15/210.69 & 0.07/0.15/0.26   & 0.23 / 0.08 / 0.17 & 3.21 & 1.24 \\
single adapter continuous & SFT & l->t & 77.68/229.62/238.58 & 0.10/0.28/0.27   & 0.19 / 0.07 / 0.14 & 3.28 & 1.28 \\
single adapter continuous & DPO & t->l & 38.87/91.74/224.39 & 0.05/0.11/0.27   & 0.28 / 0.09 / 0.18 & 4.36 & 0.60 \\
single adapter continuous & DPO & l->t & 24.42/79.56/411.97 & 0.03/0.09/0.58   & 0.28 / 0.09 / 0.19 & 4.15 & 0.77 \\
\midrule

multi adapter & SFT & t->l & 59.35/147.14/251.61 & 0.07/0.17/0.30  & 0.23 / 0.08 / 0.17 & 3.60 & 1.10 \\
multi adapter & SFT & l->t & 54.03/110.51/277.44 & 0.07/0.13/0.34 
& 0.25 / 0.08 / 0.17 & 4.26 & 0.62 \\
multi adapter & DPO & t->l & 45.26/91.99/221.61 & 0.06/0.11/0.29 &   0.27 / 0.09 / 0.18 & 4.30 & 0.51 \\
multi adapter & DPO & l->t & 34.52/91.16/222.81 & 0.04/0.11/0.29 &   0.28 / 0.09 / 0.18 & 4.32 & 0.59 \\
\midrule
HLoRA & SFT &  t->l & 53.10/68.27/103.97 & 0.06/0.07/0.13   & 0.25 / 0.08 / 0.18 & 2.67 & 1.29 \\
HLoRA & SFT &  l->t & 52.65/62.89/77.83 & 0.06/0.06/0.10 &   0.26 / 0.08 / 0.19 & 2.92 & 1.26 \\

\bottomrule
\end{tabular}%
}
\caption{
Llama-3.1-Storm-8B Model results on controlling Length and Topic
(S/N/H) refers to Short, Normal, and High for length. For metrics presented in this way we care more about the trend(whether it is increasing) rather than the exact values.
 Note that config refers to the weights of the individual attribute adapter when weighted fusion was done for the adapter fusion method and the order of training on the adapters for the rest of the rows, and that t->l refers to a topic, then length, and vice versa for l->t.
 We notice the same trend: topical summaries trained on  SFT while DPO improves the performance on topic controllability. All methods are adequately control length. 
}

\label{tab:llama_length_topic}
\end{table*}

%% file: latex/tables/length_and_specificity_mistral.tex
\begin{table*}[ht]
\setlength{\tabcolsep}{4pt}
\centering
\tiny
\resizebox{\textwidth}{!}{%
\begin{tabular}{|l|l|l|ll|l|}
\toprule
\multirow{2}{*}{} & \multirow{2}{*}{Method} & \multirow{2}{*}{Config} & \multicolumn{2}{c|}{Length Metrics(S/N/L)} & \multicolumn{1}{c|}{Specificity Metrics(N/H)} \\
\cmidrule(lr){4-5} \cmidrule(lr){6-6}
& & & Length & Comp. Ratio & Specificity \\
\midrule
zero shot &  &  & 168.76/212.59/259.22 & 0.27/0.28/0.4 & 0.76/0.74  \\
\midrule
joint training & SFT &  & 48.51/63.52/82.26 & 0.08/0.08/0.12 & 0.7/0.66  \\
joint training & DPO &  & 156.67/206.39/248.45 & 0.26/0.28/0.39 & 0.77/0.76  \\
\midrule

single adapter continuous & SFT & l->s & 36.95/53.54/84.76 & 0.06/0.07/0.12 & 0.7/0.71  \\
single adapter continuous & SFT & s->l & 54.22/60.96/74.35 & 0.08/0.08/0.11 & 0.67/0.64  \\
single adapter continuous & DPO & l->s & 125.33/188.9/230.6 & 0.2/0.26/0.37 & 0.74/0.76  \\
single adapter continuous & DPO & s->l & 161.3/205.43/234.1 & 0.26/0.28/0.37 & 0.74/0.73  \\
\midrule

multi adapter & DPO & l->s & 179.32/214.69/257.98 & 0.29/0.29/0.4 & 0.77/0.77  \\
multi adapter & DPO & s->l & 150.91/193.69/240.19 & 0.25/0.26/0.38 & 0.76/0.73  \\
multi adapter & SFT & l->s & 44.74/66.72/103.77 & 0.07/0.09/0.15 & 0.69/0.7  \\
multi adapter & SFT & s->l & 41.22/56.14/95.4 & 0.06/0.07/0.14 & 0.69/0.7  \\
\midrule

\bottomrule
\end{tabular}%
}
\caption{
Mistral-7B-Instruct-v0.3 Model results on controlling Length and Specificity.
(S/N/H) refers to Short, Normal, High for length and (N/H) refers to Normal , High for specificity. For metrics presented in this way we care more about the trend(whether it is increasing) rather than the exact values.
 Note that config refers to the order of training of attributes on the adapters and that l->s refers to length, then specificity, and vice versa for s->l.
 As expected, model is able to adequately control length but fails to do so for specificity(negligible controllability in the instance of DPO with single adapter continuously trained with on length first and then specificity. In case of not doing joint training it is marginally better to train of length first and then train on specificity.  
}

\label{tab:mistral_length_specificity}
\end{table*}

%% file: latex/tables/length_and_specificity_llama.tex
\begin{table*}[ht]
\setlength{\tabcolsep}{4pt}
\centering
\tiny
\resizebox{\textwidth}{!}{%
\begin{tabular}{|l|l|l|ll|l|}
\toprule
\multirow{2}{*}{} & \multirow{2}{*}{Method} & \multirow{2}{*}{Config} & \multicolumn{2}{c|}{Length Metrics(S/N/L)} & \multicolumn{1}{c|}{Specificity Metrics(N/H)} \\
\cmidrule(lr){4-5} \cmidrule(lr){6-6}
& & & Length & Comp. Ratio & Specificity \\
\midrule
zero shot &  &  & 53.58/110.89/218.14 & 0.08/0.15/0.34 & 0.58/0.54  \\
\midrule

joint training & SFT &  & 47.59/144.92/233.43 & 0.06/0.17/0.37 & 0.5/0.52  \\
joint training & DPO &  & 51.35/119.54/285.36 & 0.08/0.16/0.44 & 0.6/0.59  \\
\midrule
single adapter continuous & SFT & s->l & 64.58/160.95/272.88 & 0.09/0.18/0.41 & 0.51/0.51  \\
single adapter continuous & SFT & l->s & 93.25/223.67/275.93 & 0.12/0.27/0.4 & 0.51/0.52  \\
single adapter continuous & DPO & s->l & 45.05/81.57/238.33 & 0.07/0.11/0.37 & 0.58/0.57  \\
single adapter continuous & DPO & l->s & 45.83/118.53/360.97 & 0.07/0.16/0.63 & 0.62/0.6  \\
\midrule
multi adapter & SFT & s->l & 81.46/137.16/285.36 & 0.1/0.16/0.42 & 0.5/0.5  \\
multi adapter & SFT & l->s & 79.27/199.43/345.11 & 0.11/0.24/0.54 & 0.49/0.48  \\
multi adapter & DPO & s->l & 43.01/79.1/242.19 & 0.06/0.1/0.38 & 0.6/0.57  \\
multi adapter & DPO & l->s & 46.42/110.48/220.03 & 0.07/0.15/0.34 & 0.57/0.58  \\

\midrule

\bottomrule
\end{tabular}%
}
\caption{
Llama-3.1-Storm-8B results on controlling Length and Specificity.
(S/N/H) refers to Short, Normal, High for length and (N/H) refers to Normal , High for specificity. For metrics presented in this way we care more about the trend(whether it is increasing) rather than the exact values.
 Note that config refers to the order of training of attributes on the adapters and that l->s refers to length, then specificity, and vice versa for s->l.
 Length is being controlled but model is struggling to meaningfully control specificity
}

\label{tab:llama_length_specificity}
\end{table*}

%% file: latex/tables/extractiveness_and_topic_mistral.tex
\begin{table*}[ht]
\setlength{\tabcolsep}{4pt}
\centering
\tiny
\resizebox{\textwidth}{!}{%
\begin{tabular}{|l|l|l|lll|lll}
\toprule

\multirow{2}{*}{} & \multirow{2}{*}{Method} & \multirow{2}{*}{Config} & \multicolumn{3}{c|}{\centering Extractiveness Metrics(N/H/F)} & \multicolumn{3}{c}{\centering Topic Metrics} \\
\cmidrule(lr){4-7} \cmidrule(lr){7-9}

& & & Density & Coverage & Overlap Precision & Rouge(1/2/L) & G-eval Score & G-eval Std \\
\midrule
zero shot &   &  & 4.32/4.85/4.39 & 0.86/0.87/0.86 & 0.13/0.14/0.13 &  0.17 / 0.06 / 0.12 & 4.42 & 0.65 \\
\midrule

joint training & SFT &  & 4.20/5.18/2.54 & 0.86/0.84/0.85 & 0.03/0.03/0.02 &  0.22 / 0.06 / 0.15 & 2.53 & 1.28 \\
joint training & DPO &  & 3.98/4.36/3.88 & 0.85/0.87/0.86 & 0.11/0.12/0.12 &  0.19 / 0.07 / 0.13 & 4.51 & 0.60 \\

\midrule

adapter fusion & SFT & 0.67, 0.33 & 3.74/4.02/2.89 & 0.87/0.86/0.86 & 0.04/0.04/0.02 &  0.22 / 0.06 / 0.15 & 2.93 & 1.15 \\
adapter fusion & SFT & 0.5, 0.5 & 4.13/3.56/2.92 & 0.87/0.87/0.86 & 0.04/0.03/0.03 &  0.21 / 0.05 / 0.15 & 3.18 & 1.20 \\
adapter fusion & SFT & 0.33, 0.67 & 5.84/4.05/4.22 & 0.88/0.88/0.89 & 0.05/0.05/0.05 &  0.22 / 0.06 / 0.15 & 3.38 & 1.15 \\
adapter fusion & DPO & 0.67, 0.33 & 4.39/4.34/4.77 & 0.87/0.87/0.87 & 0.12/0.11/0.12 &  0.18 / 0.07 / 0.12 & 4.48 & 0.60 \\
adapter fusion & DPO & 0.5, 0.5 & 4.70/4.58/4.06 & 0.87/0.88/0.86 & 0.12/0.12/0.10 &  0.19 / 0.07 / 0.13 & 4.44 & 0.65 \\
adapter fusion & DPO & 0.33, 0.67 & 4.25/4.51/3.81 & 0.86/0.88/0.85 & 0.11/0.11/0.10 &  0.20 / 0.07 / 0.14 & 4.45 & 0.59 \\

\midrule

single adapter continuous & SFT & t->e & 3.19/4.35/3.44 & 0.83/0.86/0.86 & 0.03/0.03/0.02 &  0.21 / 0.05 / 0.14 & 2.51 & 1.23 \\
single adapter continuous & SFT & e->t & 3.22/3.36/6.80 & 0.85/0.84/0.87 & 0.02/0.01/0.02 &  0.18 / 0.04 / 0.13 & 1.86 & 1.09 \\
single adapter continuous & DPO & t->e & 4.63/4.66/4.57 & 0.87/0.88/0.87 & 0.13/0.11/0.12 &  0.19 / 0.07 / 0.13 & 4.46 & 0.61 \\
single adapter continuous & DPO & e->t & 4.14/4.15/3.92 & 0.86/0.88/0.87 & 0.09/0.07/0.07 &  0.23 / 0.08 / 0.15 & 4.44 & 0.72 \\
\midrule

multi adapter & SFT & t->e & 2.85/3.15/3.20 & 0.85/0.85/0.85 & 0.03/0.02/0.02 &  0.20 / 0.05 / 0.14 & 2.58 & 1.23 \\
multi adapter & SFT & e->t & 4.67/4.09/4.17 & 0.89/0.88/0.88 & 0.06/0.04/0.05 &  0.22 / 0.07 / 0.15 & 3.67 & 1.02 \\
multi adapter & DPO & t->e & 4.61/4.53/4.73 & 0.87/0.87/0.87 & 0.14/0.13/0.14 &  0.17 / 0.06 / 0.12 & 4.39 & 0.68 \\
multi adapter & DPO & e->t & 4.25/4.21/4.21 & 0.87/0.87/0.87 & 0.09/0.09/0.08 &  0.22 / 0.08 / 0.15 & 4.44 & 0.61 \\
\midrule

\bottomrule
\end{tabular}%
}
\caption{
Mistral-7B-Instruct-v0.3  results on controlling Topic and Extractiveness Metrics.
(N/H/F) refers to Normal, High, Fully for extractiveness. For metrics presented in this way we care more about the trend(whether it is increasing) rather than the exact values.
 Note that config refers to the weights of the individual attribute adapter when weighted fusion was done for the adapter fusion method, and order of training on the adapters for the rest of the rows, and that t->e refers to topic, then extractiveness, and vice versa for e->t.
 DPO improves performance on topic,whereas SFT degrades performance for topic. None of the method is able to make Mistral control extractiveness along with topic.
}
\label{tab:mistral_extractiveness_topic}
\end{table*}

%% file: latex/tables/extractiveness_and_topic_llama.tex
\begin{table*}[ht]
\setlength{\tabcolsep}{4pt}
\centering
\tiny
\resizebox{\textwidth}{!}{
\begin{tabular}{|l|l|l|lll|lll}
\toprule

\multirow{2}{*}{} & \multirow{2}{*}{Method} & \multirow{2}{*}{Config} & \multicolumn{3}{c|}{\centering Extractiveness Metrics(N/H/F)} & \multicolumn{3}{c}{\centering Topic Metrics} \\ 
\cmidrule(lr){4-6} \cmidrule(lr){7-9}
& & & Density & Coverage & Overlap Precision & Rouge(1/2/L) & G-eval Score & G-eval Std \\
\midrule
zero shot &   &  & 3.05/2.95/3.55 & 0.90/0.90/0.91 & 0.04/0.04/0.05 &  0.28 / 0.09 / 0.18 & 4.27 & 0.58 \\
\midrule

joint training & SFT &  & 50.49/61.78/131.10 & 0.96/0.96/0.99 & 0.12/0.18/0.29 &  0.22 / 0.08 / 0.16 & 3.71 & 1.05 \\
joint training & DPO &  & 3.02/3.13/3.87 & 0.89/0.90/0.92 & 0.04/0.04/0.05 &  0.29 / 0.10 / 0.19 & 4.33 & 0.57 \\

\midrule

adapter fusion & SFT & 0.67, 0.33 & 9.96/34.17/39.23 & 0.94/0.96/0.98 & 0.08/0.12/0.13 &  0.26 / 0.10 / 0.19 & 4.02 & 0.84 \\
adapter fusion & SFT & 0.5, 0.5 & 9.37/33.02/29.66 & 0.94/0.95/0.98 & 0.07/0.11/0.12 &  0.26 / 0.10 / 0.18 & 3.94 & 0.90 \\
adapter fusion & SFT & 0.33, 0.67 & 8.79/14.20/25.29 & 0.94/0.95/0.97 & 0.06/0.08/0.12 &  0.26 / 0.09 / 0.18 & 4.08 & 0.67 \\
adapter fusion & DPO & 0.67, 0.33 & 3.08/3.64/4.19 & 0.90/0.91/0.93 & 0.04/0.04/0.05 &  0.28 / 0.10 / 0.19 & 4.25 & 0.58 \\
adapter fusion & DPO & 0.5, 0.5 & 3.04/3.34/4.50 & 0.90/0.92/0.94 & 0.04/0.04/0.05 &  0.29 / 0.10 / 0.19 & 4.29 & 0.54 \\
adapter fusion & DPO & 0.33, 0.67 & 3.08/3.00/3.94 & 0.90/0.91/0.92 & 0.04/0.04/0.05 &  0.29 / 0.10 / 0.20 & 4.29 & 0.51 \\
\midrule
single adapter continuous & SFT & t->e & 41.79/44.36/121.42 & 0.96/0.98/0.98 & 0.10/0.12/0.22 &  0.23 / 0.08 / 0.17 & 3.29 & 1.20 \\
single adapter continuous & SFT & e->t & 76.86/89.41/144.44 & 0.97/0.98/0.99 & 0.15/0.17/0.28 &  0.21 / 0.08 / 0.15 & 3.12 & 1.31 \\
single adapter continuous & DPO & t->e & 2.85/3.30/4.26 & 0.88/0.91/0.93 & 0.04/0.05/0.06 &  0.28 / 0.09 / 0.18 & 4.39 & 0.51 \\
single adapter continuous & DPO & e->t & 3.10/4.87/5.80 & 0.90/0.93/0.95 & 0.04/0.04/0.06 &  0.30 / 0.11 / 0.20 & 4.21 & 0.65 \\
\midrule
multi adapter & SFT & t->e & 21.73/40.59/121.98 & 0.95/0.95/0.98 & 0.09/0.11/0.27 &  0.24 / 0.08 / 0.17 & 3.69 & 0.96 \\
multi adapter & SFT & e->t & 5.90/5.21/19.96 & 0.93/0.93/0.95 & 0.06/0.07/0.10 &  0.26 / 0.09 / 0.17 & 4.19 & 0.65 \\
multi adapter & DPO & t->e & 3.09/3.22/3.90 & 0.90/0.91/0.92 & 0.04/0.04/0.05 &  0.28 / 0.09 / 0.19 & 4.30 & 0.53 \\
multi adapter & DPO & e->t & 3.08/3.66/3.72 & 0.90/0.92/0.93 & 0.04/0.04/0.05 &  0.30 / 0.10 / 0.20 & 4.23 & 0.61 \\
\midrule
HLoRA & SFT & t->e & 23.96/10.35/11.25 & 0.95/0.95/0.95 & 0.05/0.04/0.05 &  0.26 / 0.09 / 0.19 & 2.50 & 1.35 \\
HLoRA & SFT & e->t & 14.26/8.93/12.38 & 0.96/0.95/0.97 & 0.04/0.04/0.04 &  0.24 / 0.07 / 0.17 & 3.41 & 1.20 \\
\bottomrule
\end{tabular}
}
\caption{
Llama-3.1-Storm-8B results on controlling Topic and Extractiveness Metrics.
(N/H/F) refers to Normal, High, Fully for extractiveness. For metrics presented in this way we care more about the trend(whether it is increasing) rather than the exact values.
 Note that config refers to the weights of the individual attribute adapter when weighted fusion was done for the adapter fusion method, and order of training on the adapters for the rest of the rows, and that t->e refers to topic, then extractiveness, and vice versa for e->t.
 We see improvements(minor in case of topic, significant for extractiveness) in controllability in over zero shot when using DPO across all the methods in this instance
}

\label{tab:llama_extractiveness_topic}

\end{table*}



%% file: latex/tables/extractiveness_and_specificity_mistral.tex
\begin{table*}[ht]
\setlength{\tabcolsep}{4pt}
\centering
\tiny
\resizebox{\textwidth}{!}{%
\begin{tabular}{|l|l|l|l|lll|}
\toprule
\multirow{2}{*}{} & \multirow{2}{*}{Method} & \multirow{2}{*}{Config} & \multicolumn{1}{c|}{Specificity Metrics(S/N/L)} & \multicolumn{3}{c|}{Extractiveness Metrics(N/H/F)} \\
\cmidrule(lr){4-4} \cmidrule(lr){5-7}
& & & Specificity & Density & Coverage & Overlap Precision \\
\midrule
zero shot &  &  & 0.54/0.55  & 3.01/3.43/5.33 & 0.9/0.92/0.92 & 0.05 / 0.05 / 0.07 \\

\midrule
zero shot &  &  & 0.78/0.76  & 4.49/4.56/4.37 & 0.86/0.87/0.86 & 0.16 / 0.13 / 0.13 \\
\midrule

joint training & SFT &  & 0.71/0.68  & 3.73/3.78/3.48 & 0.86/0.85/0.84 & 0.04 / 0.03 / 0.03 \\
joint training & DPO &  & 0.77/0.79  & 4.48/4.32/4.18 & 0.86/0.87/0.85 & 0.15 / 0.13 / 0.12 \\
\midrule

single adapter continous & SFT & e->s & 0.69/0.68  & 3.3/3.76/2.47 & 0.83/0.82/0.81 & 0.02 / 0.02 / 0.02 \\
single adapter continous & SFT & s->e & 0.67/0.66  & 2.6/2.22/2.89 & 0.84/0.82/0.84 & 0.03 / 0.02 / 0.04 \\

single adapter continuous & DPO & e->s & 0.78/0.8  & 4.5/4.37/4.29 & 0.86/0.87/0.86 & 0.16 / 0.13 / 0.12 \\
single adapter continuous & DPO & s->e & 0.73/0.73  & 4.2/4.46/4.37 & 0.86/0.87/0.85 & 0.14 / 0.14 / 0.13 \\
\midrule

multi adapter & SFT & e->s & 0.68/0.66  & 3.41/2.85/3.07 & 0.85/0.85/0.83 & 0.03 / 0.03 / 0.02 \\
multi adapter & SFT & s->e & 0.69/0.72  & 3.61/3.54/3.54 & 0.85/0.88/0.85 & 0.03 / 0.03 / 0.03 \\
multi adapter & DPO & e->s & 0.78/0.79  & 4.52/4.2/4.22 & 0.86/0.87/0.86 & 0.16 / 0.14 / 0.13 \\
multi adapter & DPO & s->e & 0.77/0.78  & 4.54/4.68/4.58 & 0.86/0.88/0.86 & 0.16 / 0.14 / 0.14 \\

\bottomrule
\end{tabular}%
}
\caption{
Mistral-7B-Instruct-v0.3 results on controlling Specificity and Extractiveness Metrics. 
(N/H/F) refers to Normal, High, Fully for extractiveness and (N/H) refers to Normal and High for Specificity. For metrics presented in this way we care more about the trend(whether it is increasing) rather than the exact values.
 Note that config refers to the order of training of attributes on the adapters, and that e->s refers to extractiveness, then specificity, and vice versa for s->e.
Mistral model is showing negligible multi-attribute controllability in the zero-shot setting, and degrades with any kind of fine-tuning 
}

\label{tab:mistral_extractiveness_specificity}
\end{table*}

%% file: latex/tables/extractiveness_and_specificity_llama.tex
\begin{table*}[ht]
\setlength{\tabcolsep}{4pt}
\centering
\tiny
\resizebox{\textwidth}{!}{%
\begin{tabular}{|l|l|l|l|lll|}
\toprule
\multirow{2}{*}{} & \multirow{2}{*}{Method} & \multirow{2}{*}{Config} & \multicolumn{1}{c|}{Specificity Metrics(N/H)} & \multicolumn{3}{c|}{Extractiveness Metrics(N/H/F)} \\
\cmidrule(lr){4-4} \cmidrule(lr){5-7}
& & & Specificity & Density & Coverage & Overlap Precision \\
\midrule
zero shot &  &  & 0.54/0.55  & 3.01/3.43/5.33 & 0.9/0.92/0.92 & 0.05 / 0.05 / 0.07 \\

\midrule

joint training & SFT &  & 0.52/0.54  & 46.57/73.99/171.14 & 0.96/0.97/0.99 & 0.12 / 0.18 / 0.29 \\
joint training & DPO &  & 0.54/0.57  & 3.08/3.68/5.86 & 0.9/0.91/0.93 & 0.06 / 0.06 / 0.07 \\
\midrule

single adapter continous & SFT & s->e & 0.52/0.53  & 72.23/104.43/106.21 & 0.97/0.96/0.99 & 0.16 / 0.15 / 0.21 \\
single adapter continous & SFT & e->s & 0.51/0.49  & 74.19/90.68/116.23 & 0.97/0.98/0.99 & 0.17 / 0.18 / 0.23 \\

single adapter continuous & DPO & s->e & 0.56/0.56  & 2.82/3.05/3.02 & 0.89/0.9/0.88 & 0.05 / 0.05 / 0.05 \\
single adapter continuous & DPO & e->s & 0.56/0.57  & 2.91/3.03/6.13 & 0.88/0.91/0.92 & 0.06 / 0.04 / 0.07 \\
\midrule

multi adapter & SFT & s->e & 0.5/0.49  & 41.49/62.98/87.63 & 0.95/0.96/0.98 & 0.12 / 0.16 / 0.22 \\
multi adapter & SFT & e->s & 0.49/0.5  & 64.2/42.48/164.11 & 0.95/0.95/0.98 & 0.18 / 0.14 / 0.31 \\
multi adapter & DPO & s->e & 0.56/0.56  & 2.89/3.81/3.62 & 0.89/0.9/0.9 & 0.05 / 0.05 / 0.06 \\
multi adapter & DPO & e->s & 0.56/0.55  & 2.94/3.29/5.23 & 0.89/0.92/0.91 & 0.05 / 0.05 / 0.06 \\

\bottomrule
\end{tabular}%
}

\caption{
Llama-3.1-Storm-8B results on controlling Specificity and Extractiveness Metrics.
(N/H/F) refers to Normal, High, Fully for extractiveness and (N/H) refers to Normal and High for Specificity. For metrics presented in this way we care more about the trend(whether it is increasing) rather than the exact values.
Note that config refers to the order of training of the adapters on the attributes, that s->e refers to specificity, then extractiveness, and vice versa for e->s. Joint training is able to control both attributes despite the fact SFT is causing large sections of the text to be copied as is. The model can control the two attributes simultaneously in the zero shot setting and is improved in the case of joint training but degrades in all other scenarios.
}
\label{tab:llama_extractiveness_specificity}
\end{table*}

%% file: latex/tables/specificity_and_topic_mistral.tex
\begin{table*}[ht]
\setlength{\tabcolsep}{4pt}
\centering
\tiny
\resizebox{\textwidth}{!}{%
\begin{tabular}{|l|l|l|l|lll|}
\toprule
\multirow{2}{*}{} & \multirow{2}{*}{Method} & \multirow{2}{*}{Config} & \multicolumn{1}{c|}{Specificity Metrics(N/H)} & \multicolumn{3}{c|}{Topic Metrics} \\
\cmidrule(lr){4-4} \cmidrule(lr){5-7}
& & & Specificity & Rouge(1/2/L) & G-eval Score & G-eval Std\\
\midrule
zero shot &   &  & 0.78/0.78 & 0.17 / 0.06 / 0.11 & 4.44 & 0.68 \\
\midrule

joint training & SFT &  & 0.66/0.71 & 0.21 / 0.05 / 0.15 & 2.61 & 1.21 \\
joint training & DPO &  & 0.78/0.79 & 0.19 / 0.06 / 0.13 & 4.46 & 0.61 \\
\midrule

single adapter continous & SFT & t->s & 0.71/0.63 & 0.20 / 0.05 / 0.14 & 2.73 & 1.20 \\
single adapter continous & SFT & s->t & 0.71/0.68 & 0.19 / 0.05 / 0.14 & 2.06 & 1.12 \\
single adapter continous & DPO & t->s & 0.78/0.78 & 0.20 / 0.07 / 0.13 & 4.50 & 0.63 \\
single adapter continous & DPO & s->t & 0.78/0.79 & 0.22 / 0.08 / 0.15 & 4.48 & 0.60 \\
\midrule

multi adapter & SFT & t->s & 0.67/0.64 & 0.20 / 0.05 / 0.14 & 2.68 & 1.23 \\
multi adapter & SFT & s->t & 0.7/0.72 & 0.22 / 0.07 / 0.15 & 3.71 & 1.02 \\
multi adapter & DPO & t->s & 0.78/0.8 & 0.17 / 0.06 / 0.11 & 4.47 & 0.65 \\
multi adapter & DPO & s->t & 0.79/0.8 & 0.22 / 0.08 / 0.14 & 4.49 & 0.59 \\

\bottomrule
\end{tabular}%
}

\caption{
Mistral-7B-Instruct-v0.3  results on controlling Specificity and Topic Metrics.
(N/H) refers to Normal, High for specificity. For metrics presented in this way we care more about the trend(whether it is increasing) rather than the exact values.
Note that config refers to the order of training of the adapters on the attributes, that s->t refers to specificity, then topic, and vice versa for t->s. DPO with either Joint training or a multi-adapter setting is the most promising avenue to control both the attribute together. We again notice SFT degrades topic performance.
}
\label{tab:mistral_topic_specificity}
\end{table*}

%% file: latex/tables/specificity_and_topic_llama.tex
\begin{table*}[ht]
\setlength{\tabcolsep}{4pt}
\centering
\tiny
\resizebox{\textwidth}{!}{%
\begin{tabular}{|l|l|l|l|lll|}
\toprule
\multirow{2}{*}{} & \multirow{2}{*}{Method} & \multirow{2}{*}{Config} & \multicolumn{1}{c|}{Specificity Metrics(N/H)} & \multicolumn{3}{c|}{Topic Metrics} \\
\cmidrule(lr){4-4} \cmidrule(lr){5-7}
& & & Specificity & Rouge(1/2/L) & G-eval Score & G-eval Std\\
\midrule
zero shot &   &  & 0.6/0.58 & 0.28 / 0.09 / 0.19 & 4.75 & 0.44 \\
\midrule 
joint training & SFT &  & 0.48/0.5 & 0.25 / 0.09 / 0.18 & 3.70 & 1.05 \\
joint training & DPO &  & 0.58/0.59 & 0.29 / 0.09 / 0.19 & 4.31 & 0.55 \\
\midrule 
single adapter continous & SFT & s->t & 0.49/0.53 & 0.20 / 0.07 / 0.15 & 3.26 & 1.25 \\
single adapter continous & SFT & t->s & 0.54/0.52 & 0.23 / 0.08 / 0.17 & 3.23 & 1.23 \\
single adapter continous & DPO & s->t & 0.64/0.65 & 0.30 / 0.10 / 0.20 & 4.29 & 0.52 \\
single adapter continous & DPO & t->s & 0.62/0.62 & 0.29 / 0.09 / 0.19 & 4.34 & 0.53 \\
\midrule

multi adapter & SFT & s->t & 0.54/0.55 & 0.27 / 0.09 / 0.18 & 4.24 & 0.64 \\
multi adapter & SFT & t->s & 0.5/0.52 & 0.24 / 0.09 / 0.17 & 3.76 & 1.02 \\
multi adapter & DPO & s->t & 0.62/0.6 & 0.29 / 0.09 / 0.19 & 4.22 & 0.64 \\
multi adapter & DPO & t->s & 0.58/0.62 & 0.28 / 0.09 / 0.19 & 4.31 & 0.56 \\

\bottomrule
\end{tabular}%
}

\caption{
Llama-3.1-Storm-8B results on controlling Specificity and Topic Metrics.
(N/H) refers to Normal and High for Specificity. For metrics presented in this way we care more about the trend(whether it is increasing) rather than the exact values.
Note that config refers to the order of training of the adapters on the attributes, that s->t refers to specificity, then topic, and vice versa for t->s. SFT shows degrading in topic controllability.
}
\label{tab:llama_topic_specificity}
\end{table*}